\documentclass[letterpaper, 10 pt, conference]{ieeeconf}
\usepackage{amsmath,amsfonts}
\usepackage{array}
\usepackage{enumerate}
\usepackage[utf8]{inputenc}
\usepackage[T1]{fontenc}
\usepackage{amsfonts}
\usepackage{amssymb}
\IEEEoverridecommandlockouts 
\usepackage{amsmath}  
\usepackage{breqn}
\usepackage[caption=false]{subfig}
\usepackage{textcomp}
\usepackage{stfloats}
\usepackage{url}
\usepackage{verbatim}
\usepackage{graphicx}
\usepackage{cite}
\usepackage{lipsum}
\usepackage{mathtools}
\usepackage[linesnumbered, ruled]{algorithm2e}

\usepackage{hyperref}
\hypersetup{
colorlinks=true,
linkcolor=blue,
filecolor=magenta,
urlcolor=blue,
}

\usepackage[font=footnotesize]{caption}

\title{\LARGE \bf

Multi-Robot Cooperative Socially-Aware Navigation Using Multi-Agent Reinforcement Learning

}

\author{Weizheng Wang$^{1}$, Le Mao$^{2}$, Ruiqi Wang$^{1}$, and Byung-Cheol Min$^{1}$ 
\thanks{This paper is based on research supported by the National Science Foundation (NSF) under Grant No. IIS-1846221. Any opinions, findings, and conclusions or recommendations expressed in this material are those of the authors and do not necessarily reflect the views of the National Science Foundation.}
\thanks{$^{1}$SMART Laboratory, Department of Computer and Information Technology, Purdue University, West Lafayette, IN, USA. {\tt\small{[wang5716 ,wang5357,minb]@purdue.edu}.}
$^{2}$College of Mechanical and Electrical Engineering, Beijing University of Chemical Technology, Beijing, China. {\tt\small{robotics.lemao@gmail.com}.} }}

\begin{document}

\maketitle

\begin{abstract}
In public spaces shared with humans, ensuring multi-robot systems navigate without collisions while respecting social norms is challenging, particularly with limited communication. Although current robot social navigation techniques leverage advances in reinforcement learning and deep learning, they frequently overlook robot dynamics in simulations, leading to a simulation-to-reality gap. In this paper, we bridge this gap by presenting a new multi-robot social navigation environment crafted using Dec-POSMDP and multi-agent reinforcement learning. Furthermore, we introduce SAMARL: a novel benchmark for cooperative multi-robot social navigation. SAMARL employs a unique spatial-temporal transformer combined with multi-agent reinforcement learning. This approach effectively captures the complex interactions between robots and humans, thus promoting cooperative tendencies in multi-robot systems. Our extensive experiments reveal that SAMARL outperforms existing baseline and ablation models in our designed environment. Demo videos for this work can be found at: \url{https://sites.google.com/view/samarl}
\label{web}

\end{abstract}

\section{Introduction}

Recent advancements in robotics and artificial intelligence have paved the way for diverse applications such as multi-robot delivery systems \cite{deliveryrobot}, collaborative exploration \cite{IJRRco-exploration}, and notably, cooperative navigation in crowded settings \cite{fan2020distributed}. Despite significant research into single-robot social navigation via both reinforcement learning (RL) \cite{kretzschmar2016socially, wang2022feedback, liu2023intention} and computation theory \cite{mavrogiannis2019multi}, the multi-robot domain remains ripe for investigation. A predominant challenge lies in defining cooperative robot behaviors while ensuring socially aware pedestrian interactions \cite{CMUsurvey}.

\begin{figure}[!t]
\centering
\includegraphics[width=0.92\columnwidth]{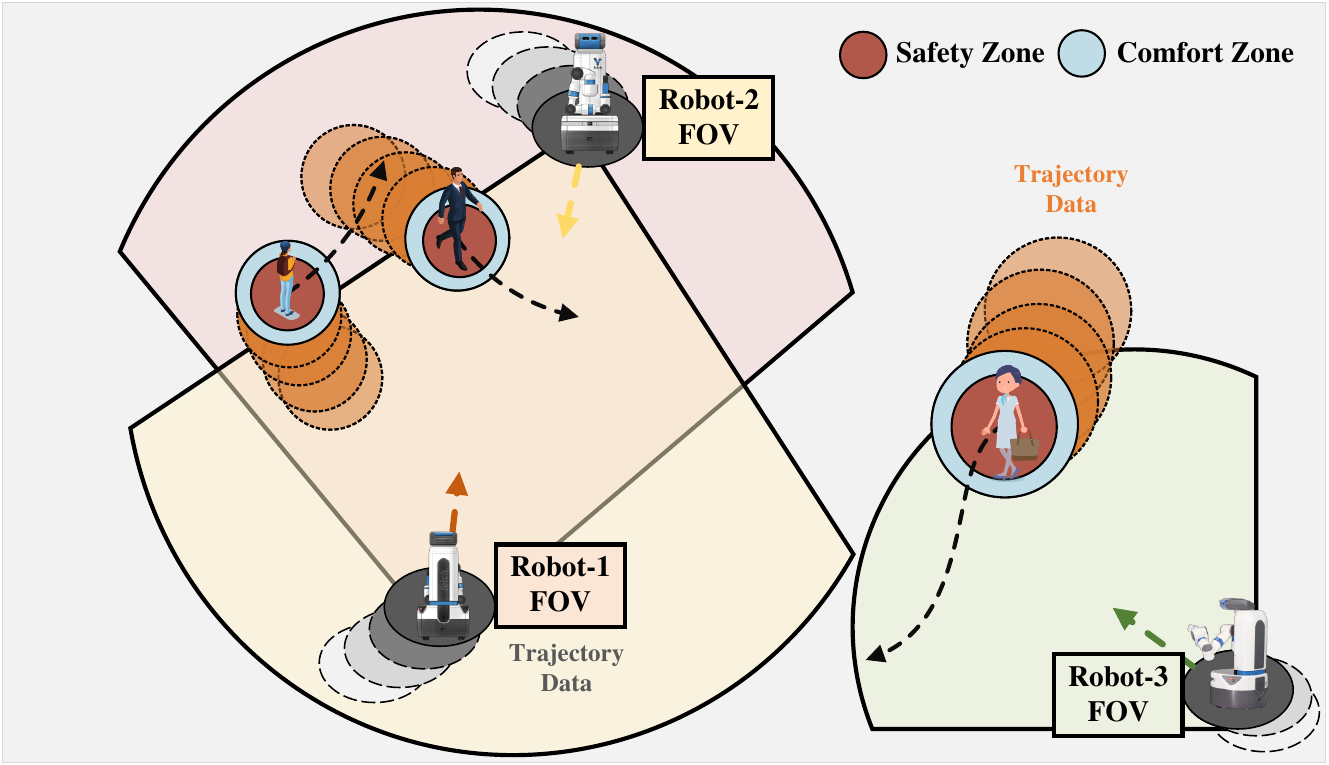}
\vspace{-4pt}
\caption{An illustration of multi-robot cooperative socially-aware navigation: social robots are engaging in cooperative navigation while maintaining a safe social distance from pedestrians.}
\vspace{-20pt}
\label{fig:F1}
\end{figure}

Central to this study is the multi-robot cooperative socially-aware navigation (MR-SAN) task (illustrated in Fig.~\ref{fig:F1}). This involves charting feasible, collision-free paths for robots in human-dense environments while respecting social norms. While real-time communication systems facilitate centralized planning for cooperative collision avoidance \cite{gasparri2017bounded}, such systems are not always viable due to potential sensor or computation delays. Therefore, we advocate for a non-communication multi-robot system that operates within the MR-SAN paradigm, where each robot makes decisions based on local observations, accounting for both human-robot interaction (HRI) and robot-robot interaction (RRI).


Existing works in both multi-robot and single-robot social navigation \cite{chen2017decentralized, liu2023intention,fan2020distributed}, despite their innovations, primarily focus on velocity-driven actions, overlooking the significance of robotic kinematics and dynamics. This omission can introduce a disconnect between simulations and real-world scenarios \cite{kadian2020sim2real}. Our paper uniquely addresses robot dynamics through non-uniform linear motion constraints and models the environment using a decentralized partially observable semi-Markov decision process (Dec-POSMDP) \cite{POSMDP}, facilitating more realistic robotic operations.


Decoupled methods \cite{cao2019dynamic,du2011robot} often separate HRI inference and path planning, leading to challenges like the freezing robot \cite{freezing} or reciprocal dance phenomena \cite{dance} in complex settings. However, contemporary learning-based approaches \cite{liu2023intention,wang2022feedback,wang2023navistar} have integrated these aspects, demonstrating success in single-robot scenarios. Addressing multi-robot navigation requires managing intricate interactions and representing cooperative paradigms, tasks that traditional single-agent RL algorithms struggle with.


To address these challenges, we present a multi-robot social navigation simulator, leveraging Dec-POSMDP and employing spatio-temporal graph (ST-graph) constructions to capture latent HRI and RRI features across both spatial and temporal dimensions. This data is processed using a hybrid spatial-temporal transformer inspired by \cite{wang2023navistar}. In order to enable effective multi-robot cooperation, we adapt a multi-agent reinforcement learning (MARL) algorithm to train joint policies and facilitate cooperative navigation behaviors. 


The main contributions of this paper are: (1) Addressing multi-robot social navigation through a Dec-POSMDP with a focus on realistic dynamics. Our method, called SAMARL, harnesses MARL \cite{yu2022surprising} to guide social robots through human-shared spaces; (2) SAMARL's deployment of a transformer-based HRI and RRI representation network derived from \cite{wang2023navistar}, to infer and align latent spatial and temporal dependencies among robots and pedestrians, facilitating social norm adherence and cooperative behaviors; (3) Demonstrating SAMARL's superior performance in both simulated and real-world contexts, outpacing current state-of-the-art (SOTA) algorithms.


    \vspace{-2mm}
\section{Related Works}
Due to the NP-hard complexity of Bayesian multistep probabilistic inference \cite{cooper1990computational}, decoupled methods have faced challenges in the long-term forecasting of crowd movements' uncertainties. In single-robot scenarios, some approaches explicitly embed constraints into planning strategies, such as using game theory \cite{turnwald2019human} or topology \cite{mavrogiannis2019multi}. Alternately, implicitly coupled HRI inference and planning align more closely with human-like thinking paradigm. For instance, in \cite{liu2021decentralized}, a ST-graph is employed to capture the spatial-temporal dependencies of environmental dynamics, and \cite{SunM-RSS-21} evaluates cooperation between humans and robots via preference distribution. Additionally, \cite{wang2022feedback} leverages human intelligence and supervision into social navigation using a preference learning framework. More recently, \cite{wang2023navistar} constructs an ST-graph based on attention correlations among spatial-temporal dimensions using a hybrid spatial-temporal transformer to enhance the effectiveness of human preference supervision. While the aforementioned approaches address the single robot socially-aware navigation task effectively, they cannot be directly applied to MR-SAN environments.


In multi-robot settings, \cite{chen2017decentralized} models pair-wise HRI for social navigation, which can extend to a multi-agent system. However, simply copying pair-wise correlation to the MR-SAN task leads to the overlook of fully and intrinsic features. For instance, robots need to maintain social distance with pedestrians under HRI and to cooperate with other robots based on RRI embedding. Despite being inspired by the impressive performance of Proximal Policy Optimization (PPO) \cite{schulman2017proximal} in various benchmarks, some works \cite{han2020cooperative, long2018towards} adapt single-agent RL algorithm PPO into multi-robot navigation tasks. But as observed in \cite{tampuu2017multiagent, yu2022surprising, kuba2022trust}, where MARL can effectively study and exhibit cooperation or competition correlations among agents compared to single-agent RL. Generally, recent MARL algorithms train agents to achieve team objectives with the centralized training and decentralized execution (CTDE) paradigm \cite{maddpg}. In this paradigm, each agent has individual actors representing personal policies, while a centralized critic is designed to capture group-wise cooperative embeddings. 

More recently, some promising works have introduced a trend that combines RL and advanced neural networks to improve action generation and feature representation. For instance, \cite{chen2021decision, mat, meng2023offline, wang2023initial} have developed transformer networks that abstract high-order distributions of semantic information into a comprehensive range of RL topics, drawing from the success of transformer networks in sequential modeling and the capability of RL in exploration. In this paper, we introduce a hybrid spatial-temporal transformer to capture long-term HRI dependencies, which is trained within a MARL training framework for multi-robot cooperative motion planning.

\section{Preliminary}
\label{sec:Preliminary}
\subsection{Kinematic \& Dynamic Configuration}
Many existing learning-based approaches \cite{fan2020distributed,chen2017decentralized,wang2023navistar,chen2020relational,chen2019crowd} assume that robotic velocities are used as actions directly, denoted as $\mathbf{a}=[v_{\rm x},v_{\rm y}]$, with the robots having infinite acceleration. However, this assumption leads to idealized robotic dynamics characterized by uniform linear motion, which can introduce discrepancies between real physical robots and simulations due to abrupt velocity changes at each time step. To address this issue, we represent robotic actions using acceleration and orientation, denoted as $\mathbf{a}^{\rm t}=[a_{\rm x}^{\rm t},a_{\rm y}^{\rm t},\theta^{\rm t}]$, assuming that robots can adjust their velocity orientation in-place at each time step. In reality, physical robots often exhibit non-uniform linear motion. To better capture real-world scenarios, we define a more comprehensive and natural kinematic configuration to model the long-term non-uniform linear motion of robots as follows:
\vspace{-1mm}
\begin{equation}
\begin{aligned}
&\mathbf{v}^{\rm{t+1}}=\mathbf{v}^{\rm{0}}+\int_{0}^{t+1} \mathbf{\overrightarrow{a}}^{\rm{t}}\rm{dt}=\mathbf{v}^{\rm t}+\mathbf{\overrightarrow{a}}^{\rm t}\Delta \rm t
\\
&\mathbf{p}^{\rm{t+1}}=\mathbf{p}^{\rm{0}}+\int_{0}^{t+1} \mathbf{v}^{\rm{t}}\rm{dt}=\mathbf{p}^{\rm t}+\mathbf{v}^{\rm t}\Delta \rm t + \frac{1}{2}\mathbf{\overrightarrow{a}}^{\rm t}\Delta \rm{t^2}
\end{aligned}
\label{eq1}
\vspace{-2mm}
\end{equation}
\noindent where $\Delta \rm t$ is the timestep of the environment, $\mathbf{\overrightarrow{a}}^{\rm t}$ is a vector quantity of acceleration at the $\rm t$-th timestep, and $\mathbf{p}$ denotes the location.

Moreover, we extend Eq. \ref{eq1} to accommodate scenarios in which certain physical robots are designed to rotate in place, for example, those equipped with Mecanum wheels \cite{wheel}. In such cases, the current velocity orientation $\theta$ becomes a part of action, enabling robots to perform in-place rotation as follows:
\begin{equation}
\begin{aligned}
&{v}^{\rm{t+1}}_{\rm x}=\Vert\mathbf{v}^{\rm{t}}\Vert_2 \cdot \cos \theta + {a}_{\rm x}^{\rm t} \Delta \rm t
\\
&{v}^{\rm{t+1}}_{\rm y}=\Vert\mathbf{v}^{\rm{t}}\Vert_2 \cdot \sin \theta + {a}_{\rm y}^{\rm t} \Delta \rm t
\end{aligned}
\end{equation}
\noindent where $\Vert \cdot \Vert_2$ represents the $\rm{L}^{2}$-norm function.

Kinematic constraints of the environments are also introduced to delineate the limitations of simulations in relation to physical robots. These constraints encompass acceleration limits, rotation restrictions, and specified turning radius conditions as follows:
\begin{equation}
\begin{aligned}
&\Vert \mathbf{\overrightarrow{a}} \Vert_2 \leq \overrightarrow{a}_{\rm max} \quad;\quad \vert \ \lim_{\rm{t \to t^+}} \theta^{\rm t}- \lim_{\rm{t \to t^-}} \theta^{\rm t} \ \vert \leq \Delta \theta_{\rm max} \quad\\
&\Delta t \cdot \Vert \mathbf{v}^{\rm t+1}\Vert_2\, > \vert \ \lim_{\rm{t \to t^-}}\theta^{\rm t+1} - \lim_{\rm{t \to t^+}}\theta^{\rm t} \ \vert
\end{aligned}
\end{equation}
\noindent where $\overrightarrow{a}_{\rm max}$ is the parameterized maximum acceleration of robots, accounting for robotic physical conditions, which is set herein as $5.0 \ m/s^2$. Also, $\Delta \theta_{\rm max}$ imposes a limit on the angular range of in-place rotation that robots can perform in one timestep (assuming $\Delta \theta_{\rm max}=\frac{\pi}{12}$ in the paper). The last equation refers to a minimum turning radius of $1.0 \ m$.

Lastly, in the environmental configuration, we assume that pedestrians have their own policies rather than being treated as simple dynamic obstacles. To maximize the performance of algorithm, we consider scenarios where robots remain unobserved by humans. This takes into account the fact that certain pedestrians, such as elderly individuals or individuals engaged in phone calls, may not promptly notice the presence of surrounding robots in real-world conditions.

\subsection{Markov Decision Process Formulation}

Taking inspiration from \cite{POSMDP,bernstein2002complexity}, we define the MR-SAN task as a  Dec-POSMDP. To accommodate the evolving landscape of social robotic operations, we employ a binary action execution strategy, as outlined in~\cite{loquercio2021learning}. This strategy consists of macro-action (MA) and local-action (LA). Initially, a MA $(\rm \hat{u})$ is generated from the multi-agent joint policy $\hat{\pi}$ at each decision-making timestep, typically associated with a global goal or waypoint. Subsequently, a set of LA $(\rm \hat{a})$ is automatically derived based on the aforementioned MA.

\begin{figure*}[!t]
\centering
\vspace{-5pt}
\includegraphics[width=0.95\linewidth]{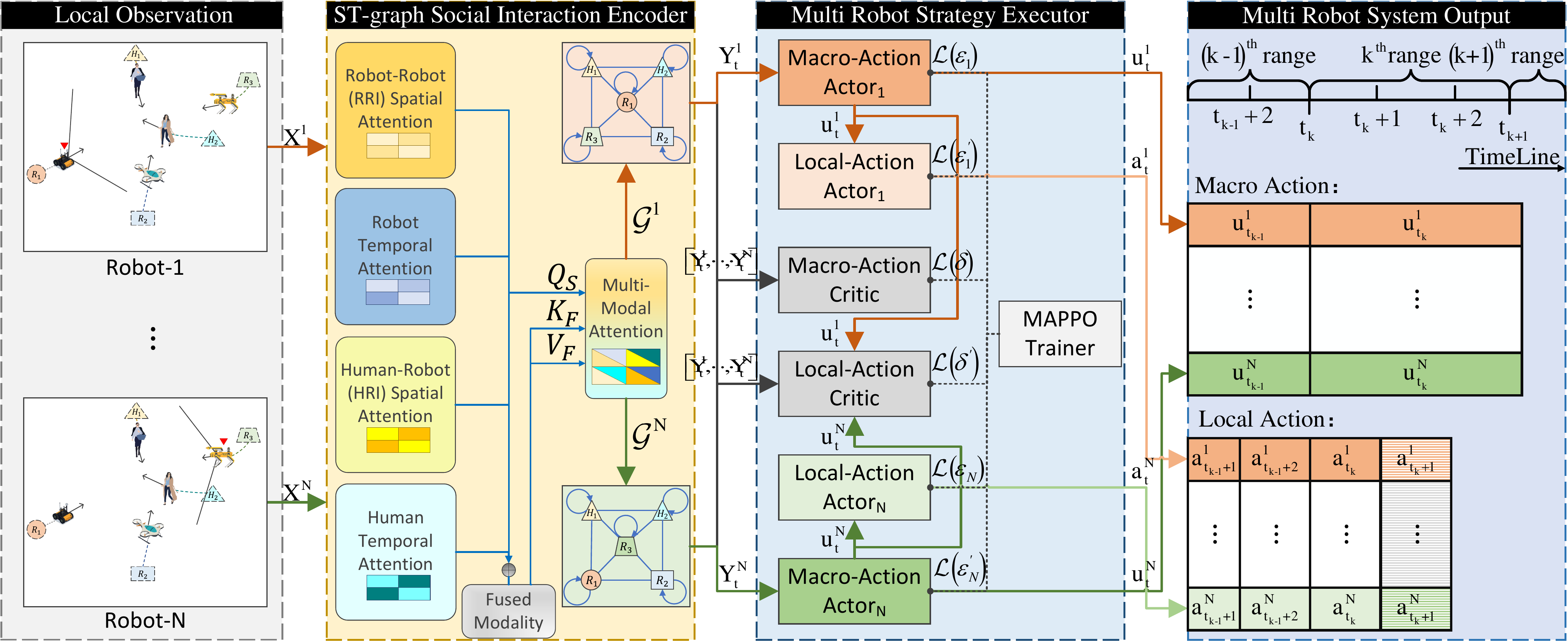}
\vspace{-5pt}
\caption{SAMARL Architecture: First, each robot feeds its individual local observations into the hybrid spatial-temporal transformer-based ST-graph social interaction encoder to create spatial-temporal state representations of HRI and RRI states. Then, the robot leverages environmental dynamics features to perform multi-robot cooperative navigation policies and adhere to social norm, using the MAPPO trainer and the social norm reward function within the multi robot strategy executor block. Finally, the generated macro-action (MA) and local-action (LA) guide the robots in the environment.}
\vspace{-15pt}
\label{fig:F2}
\end{figure*}
Firstly, the tuple $\langle \mathcal S, \mathcal U,\mathcal A,\Omega,\mathcal{O},\mathcal{P} ,\mathcal R, \mathbf{R}, \mathcal{C}, \mathcal S_0, \gamma,\rm N \rangle$ is utilized to address MR-SAN as a Dec-POSMDP. $\rm \mathbf{\hat{s}}_{\rm{t}} = [\rm \mathbf{s}^{r(1)}_{\rm{t}},\rm \mathbf{s}^{o(1)}_{\rm{t}},\cdots,\rm \mathbf{s}^{r(N)}_{\rm{t}},\rm \mathbf{s}^{o(N)}_{\rm{t}}] \in \mathcal{S}^{\rm N}$ presents that the $\rm t$-th timestep's joint state belongs to joint state space. The $\rm t$-th timestep's joint state incorporates all robots' self-states $\mathbf{s}^{r(\cdot)}_{\rm{t}}=[\rm \mathbf{s}^{r(\cdot)pu}_{\rm{t}},\rm \mathbf{s}^{r(\cdot)pr}_{\rm{t}}]$, which include individual public and privacy information, and their observation state $\mathbf{s}^{o(\cdot)}_{\rm{t}} =[\rm \mathbf{s}^{r(\cdot)pu},\cdots,\rm \mathbf{s}^{h(\cdot)pu},\cdots] \in \Omega$. This observation state can encompass other robots' or pedestrians' observable states with respect to the field of view (FOV) configuration and observable probability $\mathcal{O}$. 

For each agent (robot/human), the individual state includes an observed state and an unobserved state as $\rm \mathbf{s}_{t} = [\rm \mathbf{s}^{pu}_{\rm{t}},\rm \mathbf{s}^{pr}_{\rm{t}}]$. $ {\rm \mathbf{s}^{pu}_{\rm{t}}} = [ p_{\rm x}, p_{\rm y}, v_{\rm{x}}, v_{\rm{y}},\rho]$ is the agent's public state that can be observed by others. It includes the current position, velocity, and physical radius of the agent. $\mathbf{s}^{\rm{pr}}_{\rm{t}}=[ {g}_{\rm{x}}, {g}_{\rm{y}},v_{\rm{pref}},{\theta}]$ represents the agent's privacy state, which is not accessible to others. This state incorporates the agent's personal goal position, preferred speed \cite{chen2017decentralized}, and heading angle. 

$\mathcal{U}$ denotes the joint MA space, and $\mathcal{A}$ represents the joint LA space. At each decision-making timestep $\rm t_{k}; \rm k\in[0,\rm K]$, every robot's MA $\rm u_{t_{k}}^{i} \in \mathcal{U}$ is generated by its policy $\pi^{i}$ using the equation $\mathrm{u_{t_{k}}^{i}} \sim \pi^{\rm i}(\rm{H}_{\rm t_{k}}^{\rm i})$, and then all MAs make up a joint MA $\rm \hat{u}_{t_{k}} \in \mathcal{U}^{\rm N}$. During each time range $\rm t\in[\rm t_{k},\rm t_{k+1})$, where a total of $\rm T$ timesteps are involved from timestep $\rm t_{k}$ to timestep $\rm t_{k+1}$, each robot implements a set of LAs $[\rm a_t^{i},\cdots, \rm a_{t+T}^{i}] \sim \rm u^{i}_{t_{k}}(\rm H^{i}_{t}) \subseteq \mathcal{A}^{\rm T}$ to interact with the environment. These LAs are obtained based on the currently activated MA to form a joint LA $\rm \hat{a}_t \in \mathcal{A}^{\rm N}$. $\rm H^{i}_{t}$ is the individual history data buffer from timestep $\rm t_0$ to timestep $\rm t$. 

$\Omega$ represents the joint observation space, $\mathcal{O}^{\rm i} \rm {(s^{o(i)}|(s,a)^{i})}$ denotes the observation probability of the $i$-th agent, and $\mathcal{P}$ is the state transition probability. $\hat{\mathcal{R}}:\mathcal{S}^{\rm N}\times\mathcal{U}^{\rm N}\mapsto\mathbb{R}^{\rm N}$ is the joint MA reward function with the definition as $\hat{\mathcal{R}}(\mathrm{\hat{s},\hat{u}}) = \mathbb{E} [\textstyle\sum_{\rm t=0}^{\rm T} \gamma^{\rm t} \hat{\mathbf{R}}(\hat{\rm s}_{\rm t},\hat{\rm a}_{\rm t})|\hat{\rm a}_{\rm t} \sim \hat{\rm u}(\rm H_t) ]$. It consists of the total individual MA rewards $\mathcal{R}^{\rm i}$. On the other hand, $\hat{\mathbf{R}}:\mathcal{S}^{\rm N}\times\mathcal{A}^{\rm N}\mapsto\mathbb{R}^{\rm N}$ denotes the joint LA reward function, aggregating all agents' LA rewards $\mathbf{R}^{\rm i}$. $\mathcal{S}_{0}$ is the initial distribution, $\rm N$ is the number of social robots, and $\gamma \in [0,1]$ is a discounted factor. Refer to \cite{POSMDP} for more definitions.

In addition, the system conditional function $\mathcal{C}$ imposes constraints on the environment and multi-agent system through various termination conditions $\mathcal{C}^{\rm i}(\mathbf s_{\rm t}^{\rm i}, \mathbf a_{\rm t}^{\rm i})$. For instance, if a robot reaches its target, it will apply the brakes. Moreover, if the episode exceeds the maximum episode time $\rm t_K$ or an unfortunate collision occurs between a human-robot or robot-robot, the current episode of the environment will be terminated immediately.

In each episode, the joint initial state of the robots adheres to the initial distribution as $\hat{s}_{\rm t_0} \in \mathcal{\hat{S}}_{\rm 0}$. Initially, each robot's initial MA is obtained as $\mathrm{u_{t_{0}}^{i}}=\pi^{\rm i}(\rm{H}_{\rm t_{0}}^{\rm i})$. Then, the robots update their joint MA at each decision-making moment $\rm t_k; k\in[0,\rm K]$. Meanwhile, the robots interact directly with the environment using LAs $[\rm a_t^{i},\cdots, \rm a_{t+T}^{i}] \sim \rm u^{i}_{t_{k}}(\rm H^{i}_{t}) \subseteq \mathcal{A}^{\rm T}$, which are unfolded based on their individual current MA during each time range $\rm t\in[\rm t_{k},\rm t_{k+1})$. Subsequently, the robots collect joint reward feedback to transition to the next timestep joint state, utilizing the environmental transition probability $\mathcal{P}$. This process is either terminated or completed based on the responses of the conditional function.

\subsection{Multi Robot Socially-aware Navigation Task Statement}
\vspace{-1mm}
The objective $\mathcal{J}$ of multi robot socially-aware navigation tasks, where $\rm N$ social robots navigate alongside $\rm M$ pedestrians in an open space, can be defined as follows:
\vspace{-4pt}
\begin{equation}
\begin{aligned}
&\mathcal{J} = \mathop{\arg\min}\limits_{\forall \{\tau \in \mathcal{T}, \rm{i} \in \rm N\}}\; \textstyle\sum_{\rm{i}=1}^{\rm{N}} c_{\rm i}^{\rm t}(\tau_{\rm i}) + c_{\rm i}^{\rm s}(\tau_{\rm i}) 
\\
&=\mathop{\arg\max}\limits_{\hat{\pi}}\mathbb{E}[\textstyle\sum_{\mathrm k=0}^{\mathrm K} \gamma^{\mathrm{t_k-t_0}} \mathrm{\hat{\mathcal{R}}}(\mathrm{\hat{s}_{t_k}},\mathrm{\hat{u}_{t_k})}| (\hat{\pi},\mathrm{\hat{s}_0})]\\
&\text{s.t.} \;\; \forall \rm{i,j} \in [robot_{\rm 1},robot_{\rm N}],\forall \rm{h} \in [human_{\rm 1},human_{\rm M}]\\
&dis_{\rm i,j} > \rho_{\rm i} + \rho_{\rm j} , \, i \neq j; \;\;dis_{\rm i,h} > \rho_{\rm i} + \rho_{\rm h}; \;\; dis_{\rm i,g} < \rho_{\rm i}
\end{aligned}
\end{equation}
\noindent where $\tau$ is one of the robotic paths that belong to the clear path region $\mathcal{T}$, and $c^{\rm t}(\cdot), \text{and } c^{\rm s}(\cdot)$ are functions representing navigating time or social compliance cost. We relax the optimization of robots' cost functions as a multi-agent expectation function to address Dec-POSMDP problem. The distances between the $\rm i$-th and $\rm j$-th robots or the $\rm i$-th robot and the $\rm h$-th pedestrian are denoted by $dis_{\rm i,j}$ and $dis_{\rm i,h}$. $dis_{\rm i,g}<\rho_{\rm i}$  indicates that the $\rm i$-th robot has reached its target. In summary, we search a joint policy $\hat{\pi}=(\pi^{1},\cdots,\pi^{\rm N})$ for the multi-robot system that can maximize the expectation of the joint MA reward function. This expectation serves  as the objective of the Dec-POSMDP problem in the context of MR-SAN. This joint policy generates MAs for the multi-robots while considering the sum of total navigating cost functions for the robots, subject to certain constraints.
\section{Methodology}


\subsection{Social Interaction State Representation}
\vspace{-1mm}
We develop a social interaction state representation framework that infers environmental dynamics using an ST-graph for each robot, as depicted in Fig.~\ref{fig:F2}. This framework is based on a hybrid spatial-temporal transformer \cite{wang2023navistar}. Let each robot's self-state $\mathbf{s}^{r(\cdot)}_{\rm{t}}$ and observation state $\mathbf{s}^{o(\cdot)}_{\rm{t}}$ be contained by the robot state $\mathbf{s}^{(\cdot)}_{\rm{t}}$ as $\mathbf{s}^{(\cdot)}_{\rm{t}}=[\mathbf{s}^{r(\cdot)}_{\rm{t}},\mathbf{s}^{o(\cdot)}_{\rm{t}}]$, at $\rm t$-th timestep. The input for the $\rm i$-th robot's local FOV can be set as $\mathbf{X}^{(\rm i)}=[\mathbf{s}^{(\rm i)}_{\rm{1}},\cdots,\mathbf{s}^{(\rm i)}_{\rm{t}}]; \rm i \in [1,N]$. Moreover, we consider the MR-SAN task as a non-communication multi-agent system, where each robot must observe its surrounding environmental information independently. Due to the complexity of navigating in a multi-robot human-filled environment, relying solely on low-order HRI features may not provide sufficient performance and compliance with social norms \cite{liu2021social}. Therefore, each robot in SAMARL leverages an individual hybrid spatial-temporal transformer to infer high-order HRI and RRI dependencies in both spatial and temporal dimensions, facilitated by the construction of ST-graphs.

Initially, each robot's local input $\mathbf{X}$ is fed into spatial and temporal transformer networks in parallel to capture spatial and temporal features corresponding to the robot's local FOV. Specifically, due to the different interactive styles and targets between HRI and RRI, observed human and robot states are separately processed by independent spatial-temporal transformer blocks. For instance, SAMARL not only aims to maintain sufficient social distance between robots and pedestrians but also aim to demonstrate cooperative navigation behaviors among robots. Therefore, we divide observed human states $\mathbf{X}_{\rm H}$ and observed robot states $\mathbf{X}_{\rm R}$ to capture different interactions, resulting in RRI spatial feature $\mathbf{\hat{Y}}_{\rm RS}$, robot temporal feature $\mathbf{\hat{Y}}_{\rm RT}$, HRI spatial feature $\mathbf{\hat{Y}}_{\rm HS}$, and human temporal feature $\mathbf{\hat{Y}}_{\rm HT}$.

The spatial transformer refers to \cite{wang2023navistar} that is composed of the positional embedding layer, attention layer, graph convolution layer, and fusion gate with a residual connection. The spatial HRI features, such as correlated importance between a local agent and its neighbors with respect to relative movement intention or the static features (position and velocity) of observed agents in the FOV, can be exhibited using the attention mechanism and the graph convolution framework from spatial transformer. The successes of the vanilla transformer in language modeling, which assesses  the relative importance between pairs of words in a sentence \cite{vaswani2017attention}, is adapted into our framework to determine the importance between each pair of agents in the same timestep.
\vspace{-5pt}
\begin{equation}
    \begin{aligned}
        \operatorname{Atten}(\mathbf{X})&= \operatorname{Atten}\left(\rm \mathbf{\hat{Q}},\rm \mathbf{\hat{K}},\rm \mathbf{\hat{V}}\right)={\mathrm{softmax} (\frac{ \rm \mathbf{\hat{Q}} \rm \mathbf ({\hat{\mathbf{K}} })^{\top}}{\sqrt{ \rm{d}_{{h}}}})  \rm \mathbf{\hat{V}} }  
        \\
        \operatorname{Multi}(\mathbf{X})&=\operatorname{Multi}\left(\mathbf{\hat{Q}} ,\mathbf{\hat{K}} ,\mathbf{\hat{V}} \right)=f_{\rm{fc}}( \rm{head}_1,\cdots, \rm{head}_{\rm{h}} ) ;\\
         \rm{head}_{(\cdot)}&=\operatorname{Atten}{(\cdot)}
    \end{aligned}
\end{equation}
\noindent where $\mathbf{\hat{Q}},\mathbf{\hat{K}},\mathbf{\hat{V}}$ are the query matrix, key matrix, and value matrix of the data $\mathbf{X}$ with a dimension $\rm d_h$. $f_{\rm{fc}}(\cdot)$ denotes a fully connected layer, and the maximum head number is $\rm h$.

Similarly, a temporal transformer is also deployed to calculate the relative importance of each individual with respect to its trajectory history along the temporal dimension. This allows for the inference of agents' self-motion properties, as there is a highly relative temporal dependency in movement. The structure of the temporal transformer is similar to the spatial transformer based on \cite{wang2023navistar}, with a graph convolution layer $\rm GCN(\cdot)$ \cite{kipf2017semisupervised}.
\begin{equation}
\begin{aligned}
\{\mathbf{\hat{Y}}_{\rm{RS}}, \mathbf{\hat{Y}}_{\rm{HS}}\}&=  \rm{Trans}_{\rm{Spa}}(\rm Multi(\cdot),GCN(\cdot) |
\{\mathbf{X}_{\rm H},\mathbf{X}_{\rm R}\})\\
\{\mathbf{\hat{Y}}_{\rm{RT}}, \mathbf{\hat{Y}}_{\rm{HT}}\} &=  \rm{Trans}_{\rm{Tem}}(\rm Multi(\cdot)|\{\mathbf{X}_{\rm H},\mathbf{X}_{\rm R}\})
\end{aligned}
\end{equation}
Subsequently, a multi-modal transformer form \cite{tsai2019multimodal,wang2024husformer} is used to align heterogeneous spatial-temporal features and construct an ST-graph. Initially, the spatial and temporal features are concatenated as a fused feature. Then, the multi-modal transformer queries $\mathbf{\hat{Q}}_{\rm S}$ each single modality using keys $\mathbf{\hat{K}}_{\rm F}$ and value $\mathbf{\hat{V}}_{\rm F}$ from the fused modality, using a multi-head cross-modal attention layer. Finally, a vanilla transformer \cite{vaswani2017attention} is deployed to incorporate cross-modality features into an overall state representation $\mathbf{Y}$ as follows:
\begin{equation}
\begin{aligned}
&\rm{CMAtten}(\mathbf{X}^{\rm m}) = \rm{Multi}(\mathbf{\hat{Q}}^{\rm{m}}_{\rm{S}}, \mathbf{\hat{K}}^{\rm{}}_{\rm{F}}, \mathbf{\hat{V}}^{\rm{}}_{\rm{F}})\\
&\mathbf{{Y}}_{\rm{t}}^{\rm i} =  \rm{Trans}_{\rm{Mul}} 
(CMAtten({\mathbf{\hat{Y}}}_{\rm{t}}^{\rm i}(\cdot)|\{\rm RS,HS,RT,HT \}))
\end{aligned}
\end{equation}

Lastly, the $i$-th robot ST-graph $\mathcal{G}^{\rm i}$ is constructed to exhibit latent HRI and RRI with the graphical parameter $p_{star}^{i}$. This graph is utilized to generate robot actions and values by the policy $\pi,\pi'$ and the value function $\mathbf{V},\mathbf{V}'$.
\begin{equation}
    \mathbf Y_{\rm{t}}^{\rm i} = \mathcal{G}^{\rm i}_{\rm{t}} (\mathbf{X}^{\rm i}_{\rm t};~p_{star}^{i})
\vspace{-2mm}
\end{equation}

\begin{algorithm}[h]
 Initialize parameters $(\varepsilon, \varepsilon', \delta ,\delta', p_{star}^{i},\hat{p}_{star})$\;
\While {$step\leq step_{max}$}{
	  set data buffer $\mathcal{D}=\left\{\right\}$\;
      \For{$i=1~ to ~batch\_size$}{
           Reset the environment\; 
           Create $N$ empty caches $C = [[\ ],\dots,[\ ]]$\;
      		\For{$t_k; (k=0~ to ~K)$}{
      			\For{$all ~agents ~i=1 ~to ~N$}{
      			{\textbf{if} $\text{agent} ~i ~\text{updates} ~\text{MA}\; {{u}_{t_k}^{i}}$} in decision-making timestep $t_k$ :
                    {
                    $\mathcal{G}_{t_k}^{i} = \pi(o_{t_k}^{i};\varepsilon, p_{star}^{i}, H_{t_k}^{i})$\;
                    $\vartheta_{t_k} = \mathbf{V}(\hat{s}_{t_k}; \delta, \hat{p}_{star},H_{t_k}^{i})$\;
      		    $C^{i}+=[s_{t_k-1}^{i},o_{t_k-1}^{i},{{u}_{t_k-1}^{i}},p_{star}^{i},$ $ {\hat{p}_{star}, H_{t_k}^{i},{\mathcal{R}}_{t_k}^{i}},s_{t_k}^{i},o_{t_k}^{i}]$\;
      				  Update macro action ${{u}_{t_k}^{i}} \sim \mathcal{G}_{t_k}^{i}$\;
      			}
      		}   
        Execute $\mathrm{a}_{t}^{i}\sim \pi'(o_{t_k}^{i},{{u}_{t_k}^{i}};\varepsilon',p_{star}^{i}, H_{t_k}^{i})$\;
        $\vartheta'_{t} = \mathbf{V'}(\hat{s}_{t},\hat{u}_{t_k}; \delta', \hat{p}_{star},\hat{H}_{t})$\;        
      	}
      Compute reward and insert data into $\mathcal{D}$\;
      }
	 	Update $(\varepsilon, \varepsilon', \delta ,\delta', p_{star}^{i},\hat{p}_{star})$ on MAPPO loss\;
}      
\caption{SAMARL}
\label{alrorithm1}
\end{algorithm}
\setlength{\textfloatsep}{0.1cm}

\subsection{Multi-Agent Reinforcement Learning}

We employ one of the current SOTA MARL benchmarks, multi-agent proximal policy optimization (MAPPO) \cite{yu2022surprising}, to address the Dec-POSMDP of the MR-SAN task, following the CTDE \cite{maddpg}. During the training procedure, the algorithm utilizes information from all agents to study group-wise cooperative and competitive behaviors in the multi-robot system. However, during execution, each robot only uses individual local observations.

In SAMARL (Algorithm \ref{alrorithm1}), each robot possesses both a decentralized macro-action actor (MA-actor) and local-action actor (LA-actor). The MA-actor produces a global action $\rm u_{t_k} = [\hat{g}_x^{t_k},\hat{g}_y^{t_k}]$, i.e., the position of next waypoint, during each decision-making timestep ${\rm t_k};\rm k \in [{\rm 0},{\rm K}]$ using local observation feature $\rm \mathbf{Y}$ to guide the LA-actor in each decision-making time range $[{\rm t_k},{\rm t_{k+1}}]$. The LA-actor, in turn, generates the LA $\mathbf{a}_{\rm t}$ based on the guidance from the MA-actor and local observation features, as shown in Fig.~\ref{fig:F2}. Accordingly, with this setup, both centralized MA critic and LA critic are designed to evaluate the aforementioned actors. They do so based on joint observation features and MA information, thereby assisting the agents in presenting high-order cooperative actions from a global perspective.

SAMARL utilizes MAPPO \cite{yu2022surprising} to train each robot's actor and critic networks with the objective of demonstrating latent cooperation among the multi-robot system. In this approach, MAPPO extends PPO \cite{schulman2017proximal} to multi-agent environments through the use of Generalized Advantage Estimation (GAE) \cite{schulman2015high} and other tricks. The actor drives each robot by generating  mean and standard deviation vectors of a multivariate Gaussian distribution from local observations. The critic evaluates the value of global states to reduce variance. Finally, the parameters for actors $\varepsilon$, critics $\delta$, and the ST-graph $p_{star}$ are updated using the following loss functions $\mathcal{L}(\varepsilon)$ and $\mathcal{L}(\delta)$ as shown in Algorithm~\ref{alrorithm1}.

\setlength{\abovedisplayskip}{-10pt}
\begin{equation}
\begin{aligned}
&\mathcal{L}(\varepsilon) = \sum\limits_{\rm i=1}^{\rm N} \mathbb{E}_{\mathrm{o}\sim \Omega, \mathrm{a}\sim \mathcal{A}}[\min(\frac{\pi_{\varepsilon}(\mathrm{a}^{\mathrm{i}}|\mathrm{o}^{\mathrm{i}})}{\pi_{\varepsilon_{old}}(\mathrm{a}^{\mathrm{i}}|\mathrm{o}^{\mathrm{i}})} \mathbf{\hat{A}}^{\rm i}, clip(\frac{\pi_{\varepsilon}(\mathrm{a}^{\mathrm{i}}|\mathrm{o}^{\mathrm{i}})}{\pi_{\varepsilon_{old}}(\mathrm{a}^{\mathrm{i}}|\mathrm{o}^{\mathrm{i}})} ,\\ &1\pm\epsilon)
\mathbf{\hat{A}}^{\rm i}) + \kappa\mathbf{\hat{E}}^{\rm i}]
\\
&\mathcal{L}(\delta) = \sum\limits_{\rm i=1}^{\rm N} \mathbb{E}_{\mathrm{s}\sim \mathcal{S}} [\max((\mathbf{V}_{\delta}(\mathrm{s}^{\rm i})-\mathbf{R}^{\mathrm{i}})^{2},({clip}((\mathbf{V}_{\delta}(\mathrm{s}^{\rm i}),\\
&\mathbf{V}_{\delta_{old}}(\mathrm{s}^{\rm i})-\epsilon',\mathbf{V}_{\delta_{old}}(\mathrm{s}^{\rm i})+\epsilon')-\mathbf{R}^{\rm i})^{2}]
\end{aligned}
\end{equation}
\setlength{\abovedisplayskip}{5pt}

\noindent where $\mathbf{\hat{A}}$ is the advantage function, which is computed by GAE \cite{schulman2015high}, and $\mathbf{\hat{E}}$ is the policy entropy with an entropy coefficient hyperparameter $\kappa$.

The joint reward function $\mathbf{\hat{R}}$ is calculated by each individual reward function $\mathbf{R}(\mathbf{s}_{\rm t}^{\rm i},\mathbf{a}_{\rm t}^{\rm i})$, and the individual reward function $\mathbf{R}(\mathbf{s}_{\rm t}^{\rm i},\mathbf{a}_{\rm t}^{\rm i})$ is defined as follows:
\begin{equation}
	\mathbf{R}(\mathbf{s}_{\rm t}^{\rm i},\mathbf{a}_{\rm t}^{\rm i})=
	\begin{cases}
	5,& \text{if } \forall r\in[1,\mathrm N] \ dis_{\rm r,g}<\rho_{\rm r}\\    
     10,& \text{else } dis_{\rm i,g}<\rho_{\rm i}\\
	    -20,& \text{else } \rm{s_t} \in \mathcal{C}_{collision}(\mathbf{s}_{\rm t}^{\rm i},\mathbf{a}_{\rm t}^{\rm i}) \\
	    max(\frac{-1}{dis_{\rm i,h}},-5), & \text{else } dis_{\rm i,h} \leq 0.45 $\cite{rios2015proxemics}$ \\
        2(dis_{\rm i,g}^{\rm t-1}-dis_{\rm i,g}^{\rm t}), & \text{otherwise}
    \end{cases}
\end{equation}

\vspace{-5pt}
\section{Experiments And Results}
\subsection{Simulation Environment}

We have created a multi-robot cooperative socially-aware navigation environment, incorporating kinematic and dynamic configurations as described in Section \ref{sec:Preliminary}. Initially, we model the MR-SAN process as a Dec-POSMDP, and then address the Dec-POSMDP using a MARL algorithm with hybrid spatial-temporal transformer. In this environment, each robot's FOV is valued in a range of (0°, 360°], allowing robot to observe states of other robots and pedestrians within FOV area. Humans in the environment follow personal ORCA policies or intents. These human agents are randomly generated along a circle with radium of 5 $m$. The initial positions of the multi-robots and their respective targets are also determined based on the initial distribution $\mathcal{S}_{0}$.

\subsubsection{Baselines and Ablation Study}
We have deployed some SOTA MR-SAN algorithms as baselines for evaluation. Among these,  ORCA \cite{orca} is viewed as a representative conventional method, while CADRL \cite{chen2017decentralized} serves as an example of RL-based algorithms. Moreover, we have introduced the PPO algorithm \cite{schulman2017proximal} into the hybrid spatial temporal transformer as an ablation model, additionally, we also have replaced the transformer with SRNN \cite{chen2017decentralized}. These ablation models are referred to as SAMARL-PPO and SAMARL-SRNN, respectively, and share the same training parameters.

\subsubsection{Training Details}
All the aforementioned algorithms are trained using the same set of environmental hyperparameters. However, the network parameters are configured according to their original papers. Specifically, SAMARL, SAMRL-SRNN and SAMARL-PPO are trained over a total of $1\times10^7$ timesteps, with a learning rate of $5\times10^{-4}$ for both the actor and critic networks. Other main parameters include PPO-epoch: 5, mini-batch size: 1, gain: 0.01, clip: 0.2, entropy coefficient: 0.01. 

\subsubsection{Evaluation}
We utilize a joint success rate as our evaluation metric, ensuring that each robot can reach its individual target. Additionally, we modify the single-robot social navigation comprehensive evaluation metric $\mathrm F_{\rm SC}$ from \cite{wang2023navistar}, considering both path quality and social acceptability factors, to create the MR-SAN social score $\hat{\mathrm F}_{\mathrm{SC}}$:
\begin{equation}
    \hat{\mathrm F}_{\mathrm{SC}} = {\omega_1}{\mathrm F}_{\mathrm{SC}}^{1}+\cdots+{\omega_{\rm N}}{\mathrm F}_{\mathrm{SC}}^{\rm N}
\vspace{-4pt}
\end{equation}

\noindent where $\sum_{\rm i=1}^{\rm N}(\omega_{\rm i})=1; \omega_{\rm i} \in [0,1]$ is a set of weighted factors.

We conducted experiments with 3 robots and 10 humans, each with a FOV-360°, in 500 random cases. Moreover, we varied the parameters of robot FOV degree, human number, and robot number for more tests, as shown in Table~\ref{table:result}.

\begin{table}[h]
\vspace{-5pt}
\caption{Simulation Experiment Results\label{tab:table1}}
\vspace{-5pt}
\centering
\begin{scriptsize}
\begin{tabular}{cccccccc}
\hline & \multicolumn{3}{c}{ Success Rate } & & \multicolumn{3}{c}{MR-SAN Social Score } \\
\cline { 2 - 4 } \cline { 6 - 8 } Methods & \multicolumn{3}{c}{ FOV\&Human\&Robot } & & \multicolumn{3}{c}{ FOV\&Human\&Robot } \\
& 90° & 90° & 360° & & 90°  & 90° & 360° \\
&10\&3 & 10\&5 & 10\&5&& 10\&3& 10\&5& 10\&5\\
\hline ORCA \cite{orca}& $25$ & $21$ & $23$ & & $13$ & $8$ & $10$ \\
CADRL \cite{chen2017decentralized}& $55$ & $54$ & $63$ & & $37$ & $31$ & $33$ \\
SAMARL-PPO   & $72$ & $71$ & $72$ & & $68$ & $65$ & $67$\\
SAMARL-SRNN & $81$ & $82$ & $87$ & & $82$ & $75$ & $79$ \\
SAMARL   & $\mathbf{93}$ & $\mathbf{89}$ & $\mathbf{95}$ & & $\mathbf{94}$ & $\mathbf{85}$ & $\mathbf{91}$\\
\hline
\end{tabular}
\end{scriptsize}
\vspace{-7pt}
\label{table:result}
\end{table}

\begin{figure}[!t]
\centering
\includegraphics[width=0.75\columnwidth]{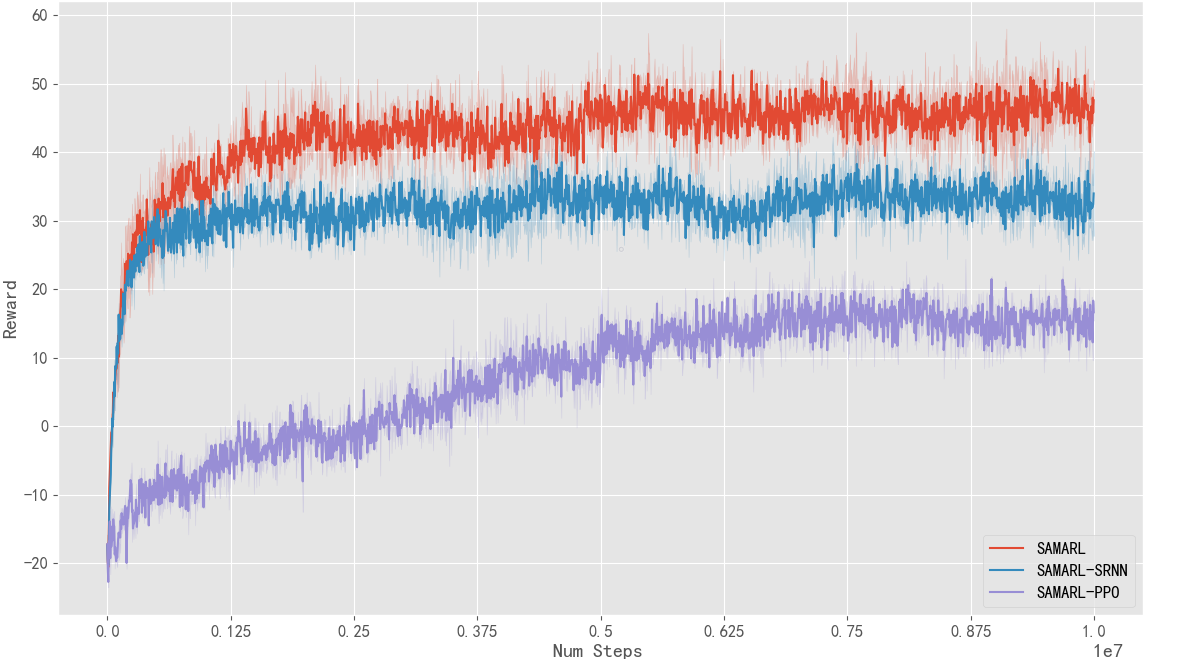}
\vspace{-4pt}
\caption{Learning curves of SAMARL and other two ablation models.}
\vspace{-1pt}
\label{fig:curve}
\end{figure}

\subsubsection{Results}

As shown in Table~\ref{table:result}, the conventional method ORCA exhibits very low performance in terms of both success rate and social score across most environmental configurations. This indicates that ORCA is ill-suited for challenging environments due to its short-sighted one-step lookahead operation. While CADRL performs better than ORCA, it still faces challenges in representing system interactions comprehensively and relies on a limited single-agent learning algorithm (Deep V-learning). The results of CADRL demonstrate that easily repeatable pair-wise interactions cannot accurately model complex HRI and RRI.

\begin{figure}[!t]
\centering
\vspace{-8pt}
\subfloat[Policy: SAMARL-PPO]{\label{fig:a} \includegraphics[width=3.8CM]{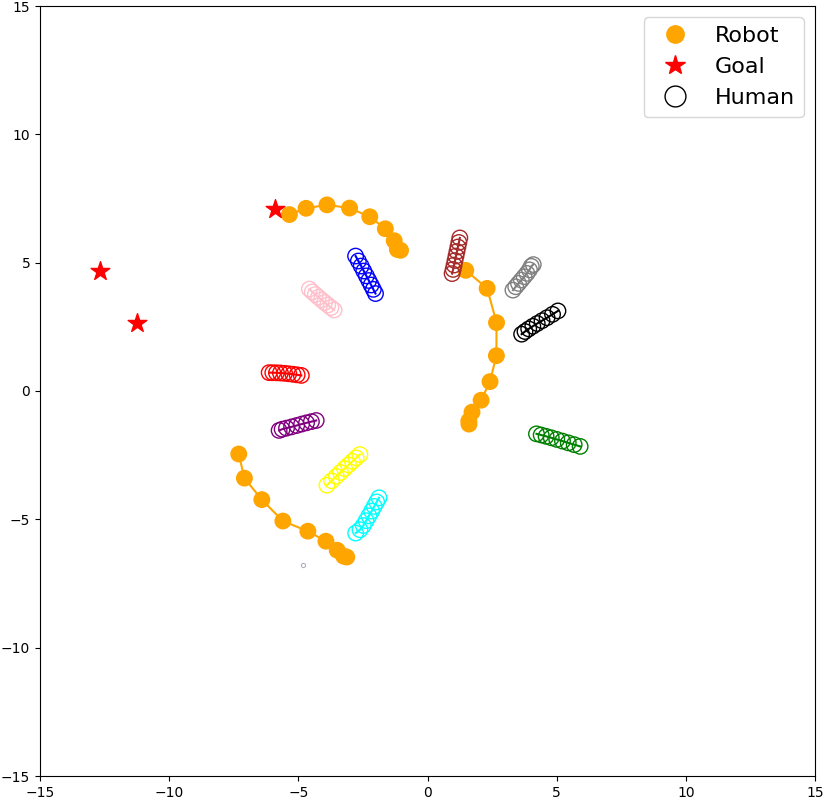}}~~
\subfloat[Policy: SAMARL]{\label{fig:b}\includegraphics[width=3.8CM]{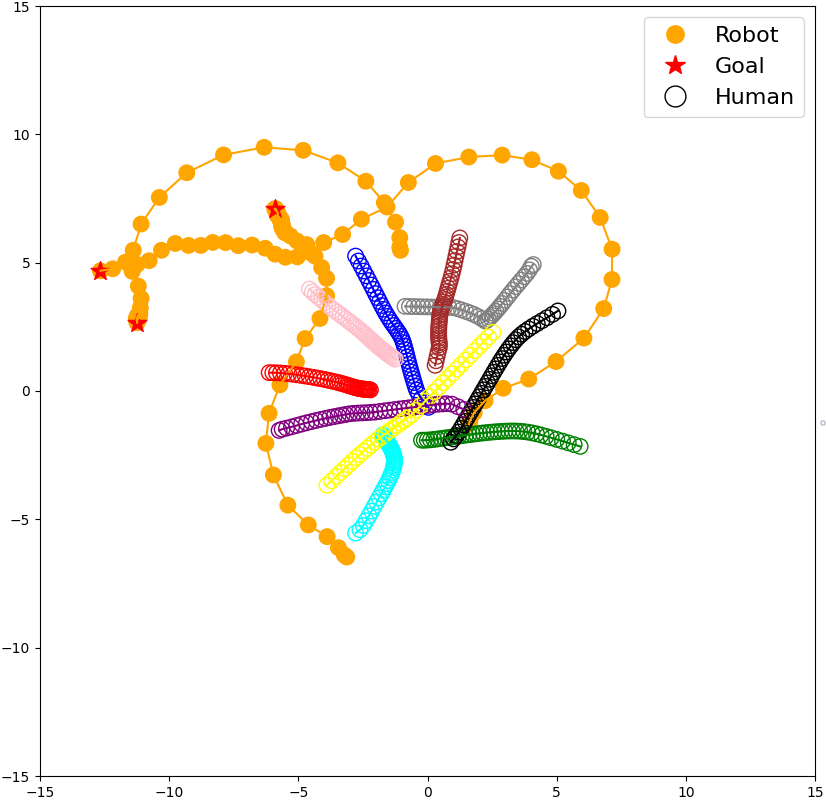}}
\vspace{-10pt}

~\subfloat[Policy: SAMARL-SRNN]{\label{fig:e}\includegraphics[width=3.8CM]{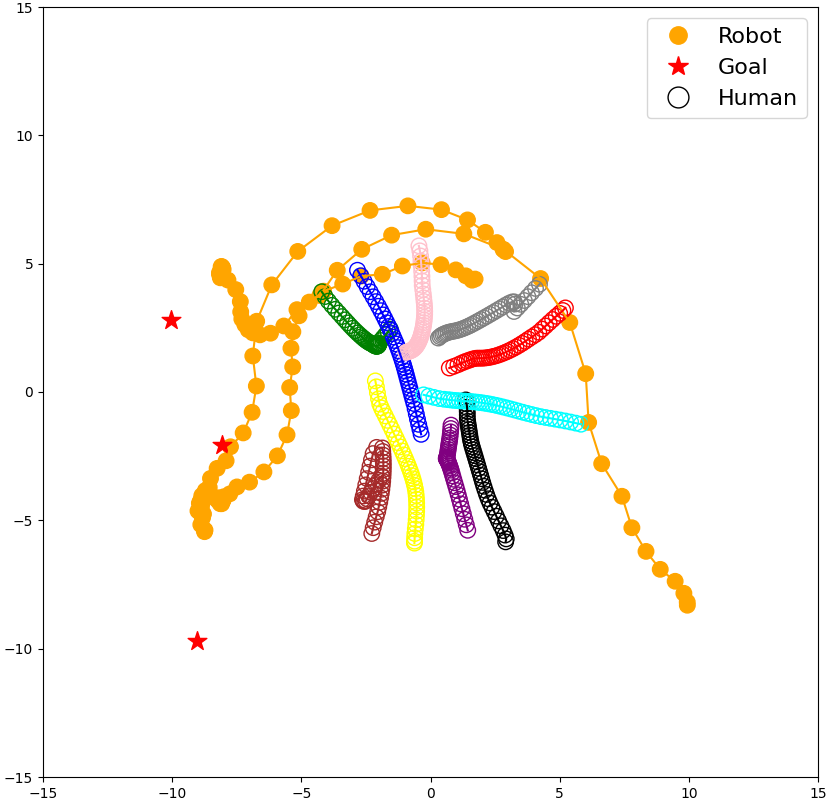}}~~
\subfloat[Policy: SAMARL]{\label{fig:f}\includegraphics[width=3.8CM]{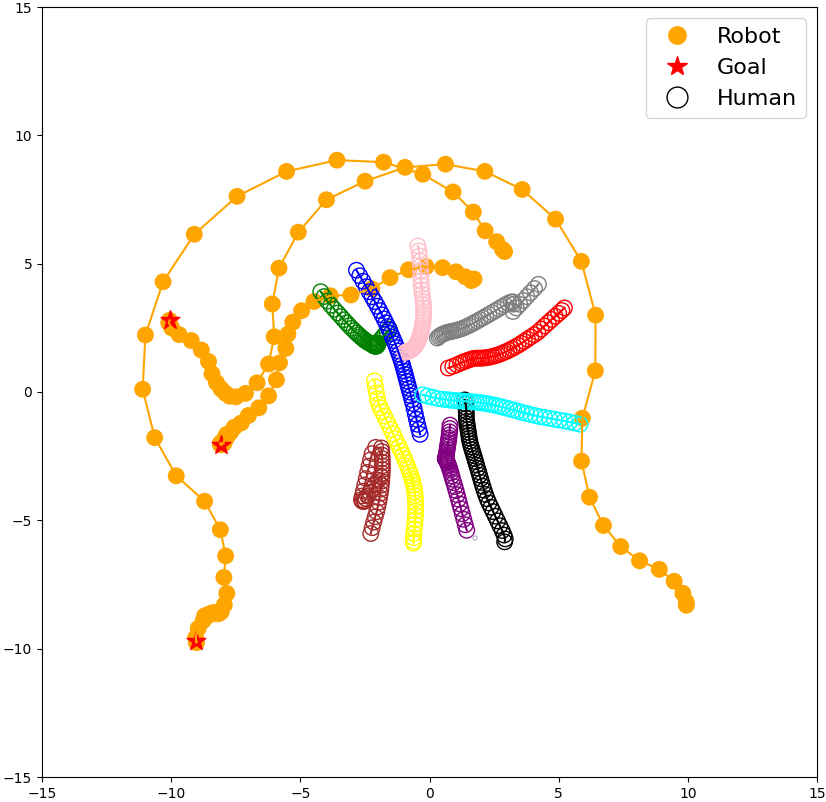}}
\vspace{-4pt}
\caption{Comparison Trajectories Visualization: the trajectories visualization of ablation models and SAMARL that are tested by the same test case.}
\vspace{-1pt}
\label{fig:traj}
\end{figure}

As shown in Table~\ref{table:result} and Fig.~\ref{fig:curve}, the deployments of new configurations generate smoother and reasonable paths than previous works for more realistic scenarios. And SAMARL demonstrates an outstanding performance compared to the other two ablation models, SAMARL-PPO and SAMARL-SRNN, in terms of evaluation metrics such as success rate, social score, and learning effectiveness in the curve. Particularly, in Fig.~\ref{fig:traj}, the first row includes the trajectory visualizations from SAMARL-PPO and SAMARL in the same test case. SAMARL-PPO resulted in an unfortunate collision between a robot and a pedestrian because single agent RL algorithms cannot reasonably capture the mapping of each agent's local observations to optimized actions. Contrary to SAMRL-PPO, the SAMRAL's training Algorithm~\ref{alrorithm1} models each agent's observation-action mapping individually, resulting in better navigation and cooperation behaviors. The second row in Fig.~\ref{fig:traj} exhibits strategic trends from SAMRL-SRNN and SAMARL, where the joint paths are too close to both humans and robots in SMARL-SRNN compared to SAMARL. Despite both SAMARL-SRNN and SAMARL utilizing the same MARL method, the ability of social interaction representation still affects performance.
\vspace{-1pt}

\vspace{-3pt}
\subsection{Real-world Experiment}
\vspace{-3pt}
We conducted a real-world experiment using two mobile robots, implementing a YOLO \cite{YOLOv8} \& DeepSORT \cite{wojke2017simple} velocity predictor with an objective tracking framework \cite{pramanik2021granulated} and a 3D localization distance estimator based on \cite{bertoni2019monoloco} as the robot perception system. This system updated observed human or robot states via a Kinect sensor. we involved 5 human participants, each walking towards individual targets while interacting with two social robots in an open space environment of $16m*16m$. The method used to generate human initial positions and targets was the same as in the simulation. As a result, our proposed method SAMARL  demonstrated promising performance with an $85.7\%$ success rate in total $14$ times real-world test cases\footnote{This physical robot experiment was reviewed and approved by the BUCT Institutional Review Board (IRB).}. Demo videos of our experiments are available on our website.


\vspace{-6pt}
\section{Conclusion}
\vspace{-5pt}
We introduce SAMARL, a benchmark for cooperative social navigation with multi-robots using MARL and transformer networks. It addresses multi-robot socially-aware navigation, incorporating HRI and RRI interactions to achieve cooperative social navigation in complex environments. Our results from simulations and real-world tests affirm its effectiveness, advancing multi-robot navigation. 



\typeout{}
\bibliography{main}
\bibliographystyle{IEEEtran}
\end{document}